\documentclass[letterpaper, 10 pt, conference]{ieeeconf}  %

\IEEEoverridecommandlockouts                              %

\usepackage{graphicx} %
\usepackage{amsmath} %
\usepackage{amssymb}  %
\usepackage{hyperref}
\usepackage{ifthen}

\usepackage{booktabs}\usepackage{multirow}
\usepackage{verbatim}

\newboolean{useJPEG}
\setboolean{useJPEG}{true}  %

\title{\LARGE \bf
Synthetically Trained Neural Networks for Learning \\ Human-Readable Plans from Real-World Demonstrations
}

\author{
	Jonathan Tremblay
	\thanks{The authors are affiliated with NVIDIA.  Email: {\tt\footnotesize \{jtremblay, thangt, styree, jkautz, sbirchfield\}@nvidia.com}}%
	\and Thang To \and 
	Artem Molchanov$^\dagger$
	\thanks{$^\dagger$Work was performed as an intern with NVIDIA.  Also affiliated with the University of Southern California.  Email: {\tt\footnotesize molchano@usc.edu}}
	\and Stephen Tyree \and Jan Kautz \and Stan Birchfield
}

\begin{document}

\maketitle
\thispagestyle{empty}
\pagestyle{empty}

%


%
\begin{abstract}
We present a system to infer and execute a human-readable program from a real-world demonstration.  
The system consists of a series of neural networks to perform perception, program generation, and program execution.  
Leveraging convolutional pose machines, the perception network reliably detects the bounding cuboids of objects in real images even when severely occluded, after training only on synthetic images using domain randomization.  
To increase the applicability of the perception network to new scenarios, the network is formulated to predict in image space rather than in world space.  
Additional networks detect relationships between objects, generate plans, and determine actions to reproduce a real-world demonstration.
The networks are trained entirely in simulation, and the system is tested in the real world on the pick-and-place problem of stacking colored cubes using a Baxter robot.\footnote{Video is at \url{https://youtu.be/B7ZT5oSnRys} .}

\end{abstract}

\section{INTRODUCTION}
\label{sec:introduction}

In order for robots to perform useful tasks in real-world settings, it must be easy to \emph{communicate} the task to the robot; this includes both the desired end result and any hints as to the best means to achieve that result.  In addition, the robot must be able to perform the task \emph{robustly} with respect to changes in the state of the world, uncertainty in sensory input, and imprecision in control output.

Teaching a robot by demonstration is a powerful approach to solve these problems.  With demonstrations, a user can communicate a task to the robot and provide clues as to how to best perform the task.  Ideally, only a single demonstration should be needed to show the robot how to do a new task.  

Unfortunately, a fundamental limitation of demonstrations is that they are concrete.  If someone pours water into a glass, the intent of the demonstration remains ambiguous.  Should the robot also pour water?  If so, then into which glass?  Should it also pour water into an adjacent mug?  When should it do so?  How much water should it pour?  What should it do if there is no water?  And so forth.  Concrete actions themselves are insufficient to answer such questions.  Rather, abstract concepts must be inferred from the actions.

We believe that language, with its ability to capture abstract universal concepts, is a natural solution to this problem of ambiguity.  By inferring a human-readable description of the task from the demonstration, such a system allows the user to debug the output and verify whether the demonstration was interpreted correctly by the system.  A human-readable description of the task can then be edited by the user to fix any errors in the interpretation.  Such a description also provides qualitative insight into the expected ability of the system to leverage previous experience on novel tasks and scenarios.

\begin{figure}
	\centering
	\ifthenelse{\boolean{useJPEG}}{
		\includegraphics[width=0.5\columnwidth]{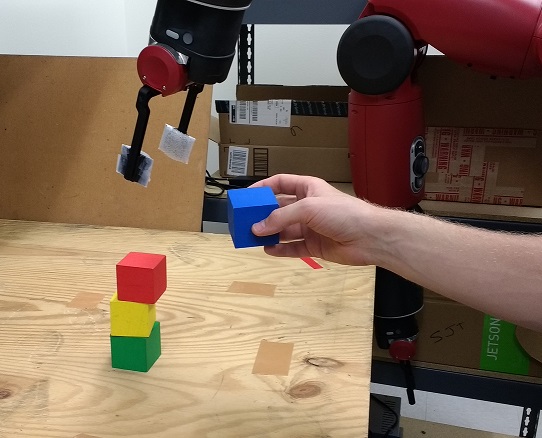}
		}
		{
		\includegraphics[width=0.5\columnwidth]{fig/overviewPic.png}
		}
	\caption{In this work, a human stacks colored cubes either vertically or in a pyramid.  A sequence of neural networks learns a human-readable program to be executed by a robot to recreate the demonstration.}
	\label{fig:pyramidIntro}
\end{figure}

In this paper we take a step in this direction by proposing a system that learns a human-readable program from a single demonstration in the real world.  The learned program can then be executed in the environment with different initial conditions.  The program is learned by a series of neural networks that are trained entirely in simulation, thus yielding inexpensive training data.  To make the problem tractable, we focus in this work on stacking colored cubes either vertically or in pyramids, as illustrated in Fig.~\ref{fig:pyramidIntro}.  Our system contains the following contributions:
\begin{itemize}
	\item The system learns from a single demonstration in the real world.  
    We believe that real-world demonstrations are more natural, 
    being applicable to a wider set of scenarios due to the reduced system complexity required, as compared 
    to AR/VR systems (e.g., \cite{duan2017arx}).
	\item The system generates human-readable plans.  As demonstrated in \cite{feniello2014iros}, this enables the resulting plan to be verified by a human user before execution.
	\item The paper introduces image-centric domain randomization for training the perception network.  In contrast with a world-centric approach (e.g., \cite{tobin2017iros}), an image-centric network makes fewer assumptions about the camera's position within the environment or the presence and visibility of fixed objects (such as a table), and is therefore portable to new situations without retraining.\footnote{Recalibrating to determine a camera's exterior orientation is arguably easier than creating a virtual environment to match a new actual environment, generating new training images, and retraining a network.}
\end{itemize}
\section{PREVIOUS WORK}
\label{sec:previous}
Our work draws inspiration from recent work on one-shot imitation learning.  Duan et al.~\cite{duan2017arx}, for example, use simulation to train a network by watching a user demonstration and replicating it with a robot. The method leverages a special neural network architecture that extensively uses soft-attention in combination with memory. During an extensive training phase in a simulated environment, the network learns to correctly repeat a demonstrated block stacking task. The complexity of the architecture, in particular the attention and memory mechanisms, support robustness when repeating the demonstration, e.g., allowing the robot to repeat a failed step.  However, the intermediate representations are not designed for interpretability. As argued in~\cite{garnello16a}, the ability to generate human interpretable representations is crucial for modularity and stronger generalization, and thus it is a main focus of our work.

Another closely related work by Denil et al.~\cite{denil17a} learns programmable agents capable of executing readable programs. They consider reinforcement learning in the context of a simulated manipulator reaching task. This work draws parallels to the third component of our architecture (the execution network), which translates a human-readable plan into a closed-loop sequence of robotic actions.  Further, our approach of decomposing the system is similar in spirit to the modular networks of \cite{devin2017icra}.

These prior approaches operate on a low-dimensional representation of the objects in the environment and train in simulation.  Like Duan et al.~\cite{duan2017arx}, we acquire a label-free low-dimensional representation of the world by leveraging simulation-to-reality transfer.  We use the simple but effective principle of domain randomization~\cite{tobin2017iros} for transferring a representation learned entirely in simulation.  This approach has been successfully applied in several robotic learning applications, including the aforementioned work in demonstration learning~\cite{duan2017arx}, as well as grasp prediction~\cite{tobin2017iros} and visuomotor control~\cite{james2017corl}.  Building on prior work, we acquire a more detailed description of the objects in a scene using object part inference inspired by Cao et al.~\cite{cao2016arx}, allowing the extraction of interpretable intermediate representations and inference of additional object parameters, such as orientation.
Further, we make predictions in image space, so that robust transfer to the real world requires only determining the camera's extrinsic parameters, rather than needing to develop a simulated world to match the real environment for training.

Related work in imitation learning trains agents via demonstrations. These methods typically focus on learning a single complex task, e.g., steering a car based on human demonstrations, instead of learning how to perform one-shot replication in a multi-task scenario, e.g., repeating a specific demonstrated block stacking sequence. Behavior cloning~\cite{pomerleau91, ross10a, ross11a} treats learning from demonstration as a supervised learning problem, teaching an agent to exactly replicate the behavior of an expert by learning a function from the observed state to the next expert action. This approach may suffer as errors accumulate in the agent's actions leading eventually to states not encountered by the expert.  Inverse reinforcement learning~\cite{Hu00, ng00irl, abbeel04} mitigates this problem by estimating a reward function to explain the behavior of the expert and training a policy with the learned reward mapping. It typically requires running an expensive reinforcement learning training step in the inner loop of optimization or, alternatively, applying generative adversarial networks~\cite{ho16, hausmann17} or guided cost learning~\cite{finn16}.
The conjunction of language and vision for environment understanding has a long history.  Early work by Winograd~\cite{winograd1971thesis} explored the use of language for a human to guide and interpret interactions between a computerized agent and a simulated 3D environment.  Models can be learned to perform automatic image captioning~\cite{vinyals2015cvpr}, video captioning~\cite{shen2017arx}, visual question answering~\cite{gupta2017arx}, and understanding of and reasoning about visual relationships~\cite{karpathy2015cvpr,peyre2017arx,johnson2017cvpr}---all interacting with a visual scene in natural language.

Recent work has studied the grounding of natural language instructions in the context of robotics~\cite{paul2017ijcai}.  Natural language utterances and the robot's visual observations are grounded in a symbolic space and the associated grounding graph, allowing the robot to infer the specific actions required to follow a subsequent verbal command.

Neural task programming (NTP) \cite{xu2017:ntp}, a concurrent work to ours, achieves one-shot imitation learning with an online hierarchical task decomposition.
An RNN-based NTP model processes a demonstration to predict the next sub-task (e.g., ``pick-and-place'') and a relevant sub-sequence of the demonstration, which are recursively input to the NTP model.
The base of the hierarchy is made up of primitive sub-tasks (e.g., ``grip'', ``move'', or ``release''), and recursive sub-task prediction is made with the current observed state as input, allowing closed loop control.
Like our work, the NTP model provides a human-readable program, but unlike our approach the NTP program is produced \emph{during} execution, not before.

\section{METHOD} 
\label{sec:method}
\begin{figure*}
	\centering
		\includegraphics[width=1.8\columnwidth]{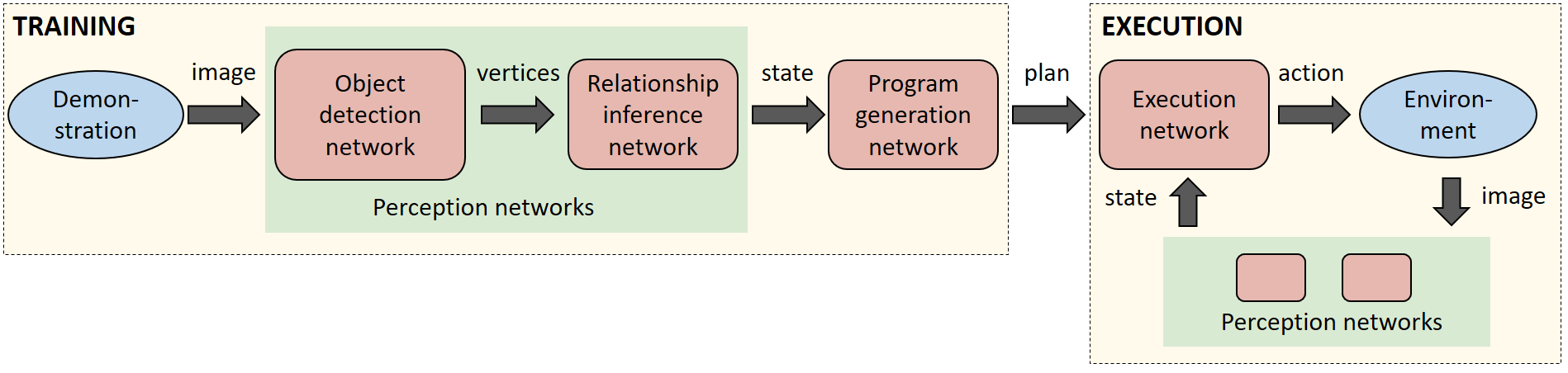}
	\caption{System overview.  As a user performs a demonstration in the real world, the perception networks detect the pose of the objects in the image domain, as well as their relationships.  These percepts are fed to a network that generates a plan, which can then be implemented by the execution network.}
	\label{fig:systemOverview}
\end{figure*}

An overview of our system is shown in Fig.~\ref{fig:systemOverview}.  A camera acquires a live video feed of a scene, and the positions and relationships of objects in the scene are inferred in real time by a pair of neural networks.  The resulting percepts are fed to another network that generates a plan to explain how to recreate those percepts.  Finally, an execution network reads the plan and generates actions for the robot, taking into account the current state of the world in order to ensure robustness to external disturbances.

\subsection{Perception networks} %
Given a single image, our perception networks infer the locations of objects in the scene and their
relationships.  These networks perform object detection with pose estimation, as well as relationship inference.

\subsubsection{Image-centric domain randomization}

Each object of interest in this work is modeled by its bounding cuboid consisting of up to seven visible vertices and one hidden vertex. Rather than directly mapping from images to object world coordinates, the network outputs values in the image coordinate system.  This makes the system robust to changes in camera position and orientation, as well as making it independent of the contents of the background of the scene (e.g., it does not need to see a table of a particular size, as in~\cite{tobin2017iros}).  Using image coordinates also makes the results easier to visualize.

\begin{figure}
	\centering
		\includegraphics[width=0.8\columnwidth]{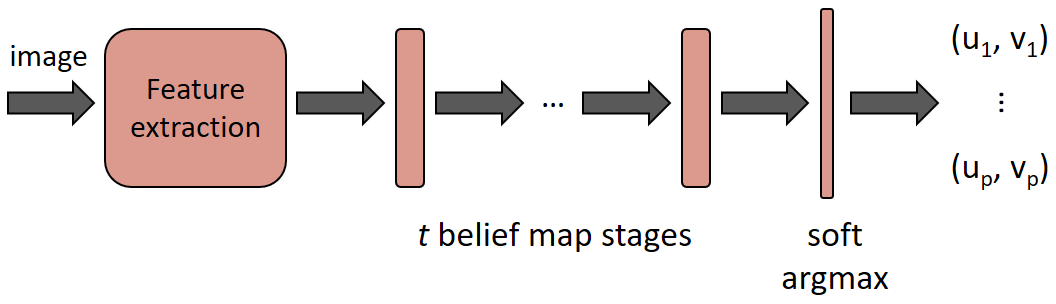}
	\caption{Architecture of the perception network for detecting a single object from a $400 \times 400 \times 3$ RGB input image.  Features are extracted using the first ten layers of VGG-19, followed by a series of belief map tensors that leverage increasingly larger amounts of context.  Each $50 \times 50 \times p$ belief map tensor contains one belief map for each of the $p=7$ vertices of the bounding cuboid.  Finally, a soft argmax operation outputs the $(u,v)$ image coordinates of each vertex.  Multiple objects are detected by running multiple networks simultaneously.} %
	\label{fig:perceptionArchitecture}
\end{figure}

\begin{figure*}
	\centering
	\begin{tabular}{ccccc}
	\ifthenelse{\boolean{useJPEG}}{
		\includegraphics[trim={12mm 12mm 12mm 12mm},clip,width=.25\columnwidth]{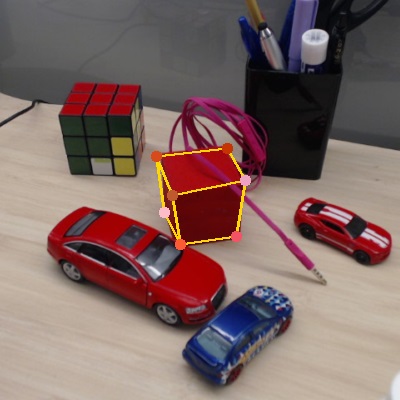} &
		\includegraphics[width=.25\columnwidth]{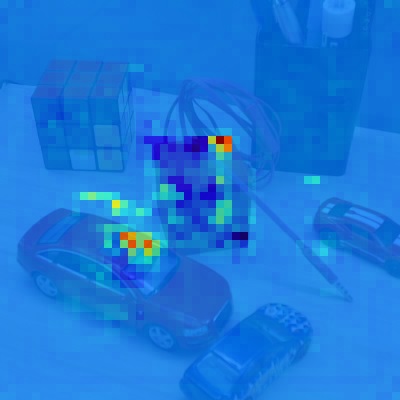} &
		\includegraphics[width=.25\columnwidth]{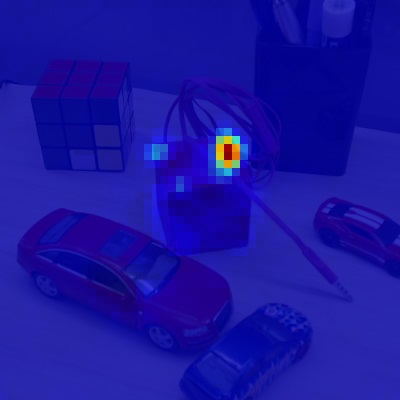} &
		\includegraphics[width=.25\columnwidth]{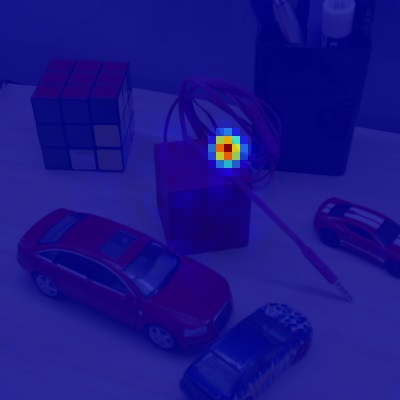} &
		\includegraphics[width=.25\columnwidth]{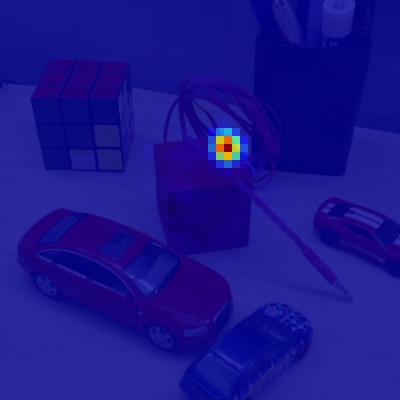} \\
		}
		{
		\includegraphics[trim={12mm 12mm 12mm 12mm},clip,width=.25\columnwidth]{fig/out_00068.png} &
		\includegraphics[width=.25\columnwidth]{fig/output0.png} &
		\includegraphics[width=.25\columnwidth]{fig/output1.png} &
		\includegraphics[width=.25\columnwidth]{fig/output2.png} &
		\includegraphics[width=.25\columnwidth]{fig/output5.png} \\
		}	
		image & stage 1 & stage 2 & stage 3 & stage 6
		\end{tabular}
	\caption{An image with detected vertices overlaid, and belief maps $B_j^t$ for stages $t \in \{1,2,3,6\}$.  Ambiguity in the early stages due to small receptive fields is resolved in later stages by taking into account more context.}
	\label{fig:beliefMaps}
\end{figure*}

The network architecture is illustrated in Fig.~\ref{fig:perceptionArchitecture}.  Feature extraction consists of the first ten layers of VGG-19 \cite{simonyan2015iclr}, pre-trained on Imagenet \cite{{deng2009in}}.  
Inspired by convolutional pose machines \cite{wei2016arx,cao2016arx}, the output of these layers are fed into a series of $t$ stages.  Each stage is a series of convolutional / ReLU layers (the same number of layers, type of layers, stride, and padding as in \cite{cao2016arx}) with weights that are learned during training.   These stages output belief maps for each vertex, with increasingly larger receptive fields to capture more of the surrounding context and resolve ambiguity. These stages are depicted in Fig.~\ref{fig:beliefMaps} for the prediction of a single vertex.

We use the $L_2$ loss between the predicted belief maps and the ground truth of the training data.  Specifically, the loss function for stage $i$ is
\begin{equation}
f_i = \sum_{j = 1}^p\left[ \sum_{u,v}  \left\| B_j^i(u,v) - {\hat B}_j(u,v) \right\|_2  \right],
\label{eq:loss}
\end{equation}
where ${\hat B}_j$ is a Gaussian-smoothed ($\sigma=3.4$~pixels in the $400 \times 400$ image) ground truth belief map for vertex $j \in \{1..p\}$ and
$B_j^i$ is the neural network output for the belief map at stage $i \in \{1..t\}$ for 
vertex $j$, where $p=7$ is the number of vertices estimated.  The total loss is the sum of the losses of the individual stages:  $f = \sum_{i=1}^t f_i$.  
This approach of applying the loss at each stage (also known as \emph{intermediate supervision} \cite{wei2016arx}) avoids the vanishing gradient problem by restoring the gradients at each stage.

Each belief map is treated as a probability mass function for the location of a vertex in the image domain.  To obtain image coordinates, soft argmax is applied along the rows and columns of the final belief map.

\begin{figure}
	\centering
	\begin{tabular}{cc}
	\ifthenelse{\boolean{useJPEG}}{
		\includegraphics[width=0.35\columnwidth]{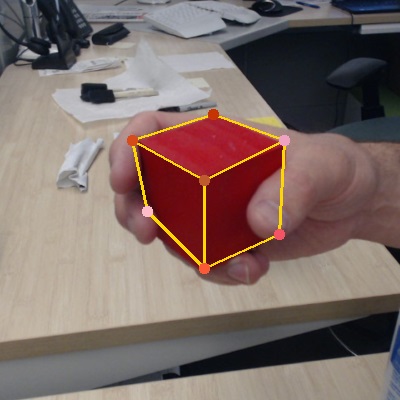} & 
		\includegraphics[width=0.35\columnwidth]{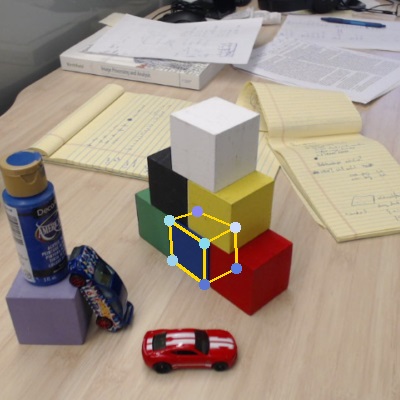} 
		}
		{
		\includegraphics[width=0.35\columnwidth]{fig/00521.png} & 
		\includegraphics[width=0.35\columnwidth]{fig/cubestack_blueocc.png} 
		}	
	\end{tabular}
	\caption{Examples of object detection from image-centric domain randomization, showing the seven detected vertices.  Shown are detections of a red cube (left) and blue cube (right), even when the latter is severely occluded.}
	\label{fig:cubestackblueocc}
\end{figure}

Examples of cuboid object detection are shown in Fig.~\ref{fig:cubestackblueocc}.  
The image-centric representation makes it easy to visualize whether the object has been detected accurately.  In addition, detecting the individual vertices yields a rich representation to facilitate estimation of the full pose of the object.  By training on instances of occluded objects, the network learns to detect the object even when it is severely occluded, as shown in the figure.

The network was trained entirely on simulated data.\footnote{Synthetic image data were generated using the NVIDIA deep learning dataset synthesizer (NDDS), \url{https://github.com/NVIDIA/Dataset_Synthesizer/} .}  Colored cubes ($5$~cm per side) were placed within a virtual environment consisting of various distractor objects in front of a background.  Images were generated by randomly varying distractors, texture, background, camera pose, lighting, and noise.  For each color, $12$k synthetic images were collected for training. 
\subsubsection{Relationship inference}

After objects have been detected, their relationships are inferred.  
This is accomplished via a fully connected neural network with 28 inputs,\footnote{That is, 2 objects, 7 vertices per object, and $(u,v)$ coordinates for each vertex.} two outputs, and three hidden layers with $100$ units in each.
The inputs to the network are the image coordinates of the vertices of two detected cuboids, and the output is an unnormalized distribution over the symbols {\sc Above} and {\sc Left}; values below a threshold yield a third symbol, {\sc None}. This set is rich enough to allow the system to build not only stacks of cubes but also more complicated structures such as pyramids.  With $n$ detected objects, the pairwise network is run $n(n-1)$ times to generate the full state of the scene from a single image. The state is represented as an $n \times n \times k$ tensor (with the diagonal ignored), where $k=3$ is the number of relationships considered.\footnote{Although such a representation could lead to self-contradiction (e.g., blue is above red, and red is simultaneously above blue), it also allows the possibility of discarding inferences with low confidence.}  Note that in the case of a pyramid, an object is above two other objects, which is manifested by two elements of the tensor having roughly equal weight.\footnote{To improve robustness, we trained a separate relationship inference network to handle this case.}

The relationship network was trained on the vertex coordinates of the simulated data mentioned above with a cross-entropy loss function.  To make the network more robust to real-world noisy inputs, these coordinates were perturbed by randomly swapping vertices (with probability $1\%$) to another vertex within the same cube and adding Gaussian noise to the observed vertex coordinates.  Moreover, occluded vertices were randomly relocated (with probability $50\%$) using a uniform distribution within the convex hull of the occluder.
\subsection{Program generation network}

The purpose of the system is to learn a human-readable program from a real-world demonstration.  While the camera watches the scene, an agent (such as a person) moves the blocks on the table to stack them vertically or in a pyramid structure.  Multiple stacks are allowed, in which case the order of operations between the different stacks is arbitrary.

As the demonstration is being performed, the perception network detects the objects and their relationships.  Once the demonstration is complete, the state tensor from the relationship inference is thresholded (at $0.5$) to yield a set of discrete relationships between the objects.  This tensor is sent to a program generation network which outputs a human-readable plan to execute. 

Our framework assumes that the demonstration involves a sequence of pick-and-place operations.  Each step of the program can therefore be represented as a $2 \times (n+1)$ binary array of values indicating which of the $n$ objects (or none) is the source (the object to be picked), and which of the $n$ objects (or none) is the target (the object upon which the picked object is to be placed).  Since with $n$ objects, there are at most $n-1$ steps in the program, the output of the program generation network is a $2 \times (n+1) \times (n-1)$ tensor of floating-point values which are converted to binary by applying argmax.  The resulting binary tensor is then trivially converted to a human-readable program.  For example, the following binary tensor
\begin{quote}
\begin{center}
\begin{small}
\begin{tabular}{rr|cccc}
\toprule
& & \multicolumn{4}{c}{\emph{place}} \\
& & red & green & blue & yellow \\
\midrule 
\multirow{4}{*}{\rotatebox[origin=c]{90}{\emph{pick}}}
& red & - & 10 & 00 & 00 \\
& green & 00 & - & 00 & 00 \\
& blue & 01 & 00 & - & 00 \\
& yellow & 00 & 00 & 00 & - \\
\bottomrule
\end{tabular} 
\end{small}
\end{center}
\end{quote}
would be translated as follows:
\begin{quote}
\begin{center}
\emph{``Place the red cube on the green cube, then place the blue cube on the red cube.''}
\end{center}
\end{quote}
The network is implemented as a fully connected double-headed neural network with seven layers (one input, one output, and five hidden) in each of the two paths, with $1024$ units in each hidden layer.
The network is trained using an MSE loss function with data from simulated scenes generated by enumerating possible states and corresponding plans.

\subsection{Execution network}

Once a program has been generated, it could be executed in an open-loop fashion by sequentially performing each step.  However, in order to allow recovery from manipulation mistakes, or to handle external disturbances, a network is used to guide execution. This execution network is fully connected with two inputs, one output, and five hidden layers with $128$ units in each.  It takes as input the program (represented as a $2 \times (n+1) \times (n-1)$ tensor) and the current state of the scene (represented as an $n \times n \times k$ tensor), and it returns the next action to take.  This action is represented by a $2 \times n$ array of values indicating the source and target objects, along with a $1 \times (k-1)$ array indicating the relationship to achieve (e.g., place one object on top of another).  The execution network is trained on synthetic data generated by randomly sampling among the possible programs and states, with an MSE loss function.

\subsection{Entire system}

The different networks presented in this section are linked together to create a combined system for learning by demonstration.  The human demonstrates a task once in the real world, from which the system infers a program.  Once the demonstration is complete, the objects may be shuffled on the working surface to change the initial conditions, and the robot can then execute the program in a closed loop manner, correcting mistakes when operations fail or perturbations are introduced.

\section{EXPERIMENTAL RESULTS}\label{sec:results}
In this section we report tests of the individual components of the system, followed by tests of the entire end-to-end system.

\subsection{Cube detection}

We begin by assessing how accurately the object detection network, trained using image-centric domain randomization, can detect the location and pose of colored cubes in real images.
We collected a dataset of images of cubes in various arrangements on a table.
Images were captured to test the effects of (a) cube color, (b) position of the cube on the table, (c) height of the cube above the table, (d) presence of other distractor cubes, (e) occlusion, and (f) position and orientation of the camera with respect to the cube.
The images were manually labeled to indicate the vertex locations, and the output of the detection network was compared with these ground truth labels.

Several measures of accuracy could be used.  For an interpretable measure of the accuracy with which points can be detected, either the mean absolute error (MAE) or root mean square (RMS) error are possible.  For determining how likely a point is detected within some radius of the ground truth, variants of the PCP metric \cite{ferrari2008cvpr} such as PCK \cite{yang2013pami} or PCKh \cite{andriluka2014cvpr} are common choices.\footnote{In the context of human pose estimation, PCP is the percentage of correctly localized \emph{parts} with respect to the length of the ground truth segment; PCK is the percentage of correctly localized \emph{keypoints} with respect to the size of the ground truth bounding box of the person; PCKh is like PCK except with respect to the size of the head rather than the person to reduce the effects of body pose.  In the latter two cases, the size of the bounding box is the maximum of length and width.}  

In this work, to measure the expected error in pixels, we use the MAE rather than RMS since it is less sensitive to outliers.  To determine the probability that a vertex will be detected correctly, we use PCKh but modified to normalize the error by the size of the cube in the image (the square root of the number of pixels in the convex hull) instead of the size of the human head.  We also introduce a third measure, MAEc, which is the MAE of the vertices that are correctly detected according to PCKh (thus ignoring outliers).  All three measures can be expressed as 
\begin{equation}
\frac{\sum_{i=1}^N \phi_i \delta_i}{\sum_{i=1}^N \psi_i},
\label{eq:phidelta}
\end{equation}
where $N$ is the total number of vertices in all the images being used for evaluation, and $\delta_i$, $\phi_i$, and $\psi_i$ are given by:
\begin{center}
\begin{tabular}{rccc}
\toprule
& $\delta_i$ & $\phi_i$ & $\psi_i$ \\
\midrule
MAE & $d_i$ & $1$ & $1$ \\
PCKh & $1$ & $c_i$ & $1$ \\
MAEc & $d_i$ & $c_i$ & $c_i$ \\
\bottomrule
\end{tabular}
\end{center}
where $c_i=1$ if $(d_i/\sqrt{A} \le \epsilon)$ or 0 otherwise, $d_i=\sqrt{(u_i-{\hat u}_i)^2+(v_i-{\hat v}_i)^2}$ is the Euclidean distance in pixel space between the detected vertex and the ground truth, $A$ is the number of pixels in the convex hull of the cube, and $\epsilon$ is a threshold.

\begin{figure}
	\centering
	\begin{tabular}{p{0.88in}p{0.88in}p{0.88in}}
	\ifthenelse{\boolean{useJPEG}}{
			\includegraphics[width=0.29\columnwidth]{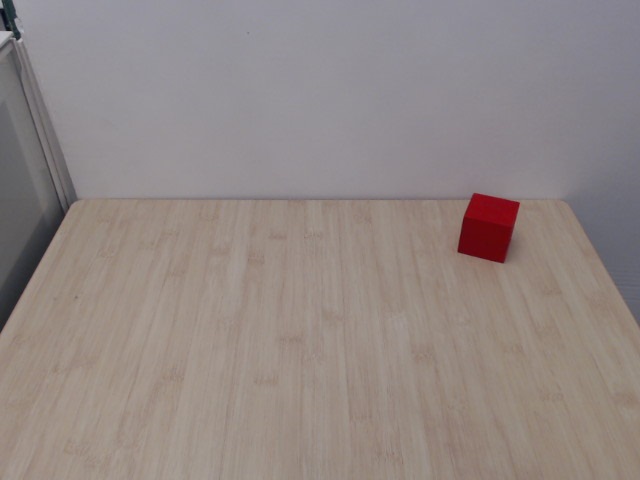} & 
			\includegraphics[width=0.29\columnwidth]{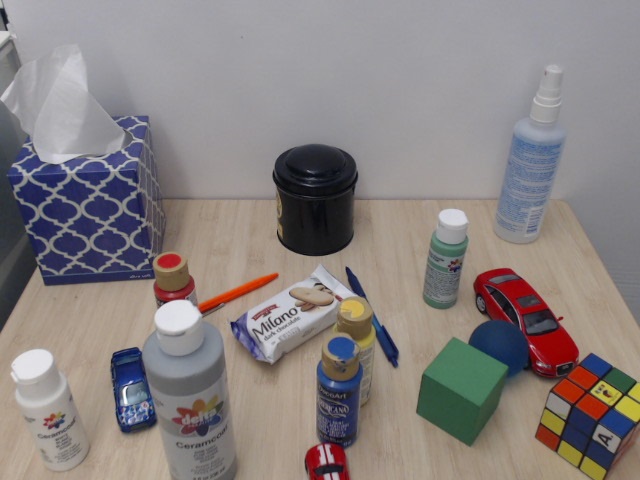} & 
			\includegraphics[width=0.29\columnwidth]{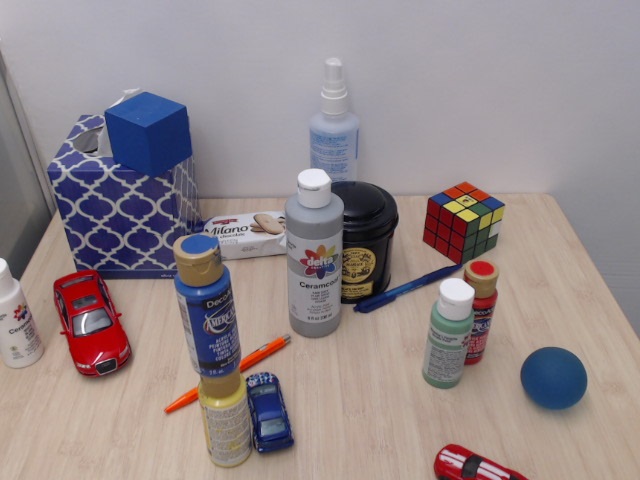} \\
			\includegraphics[width=0.29\columnwidth]{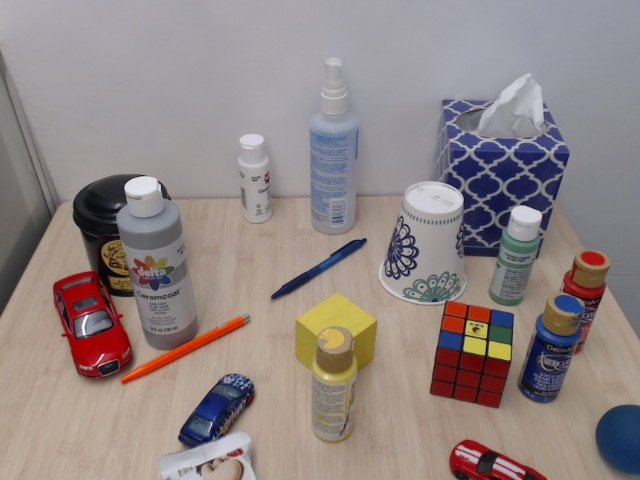} &
			\includegraphics[width=0.29\columnwidth]{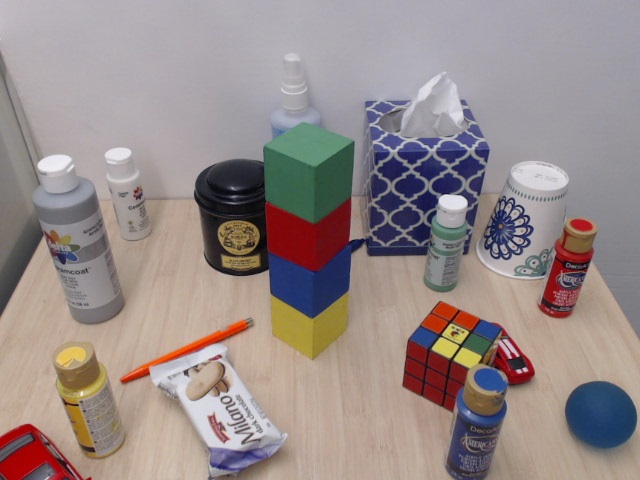} &
			\includegraphics[width=0.29\columnwidth]{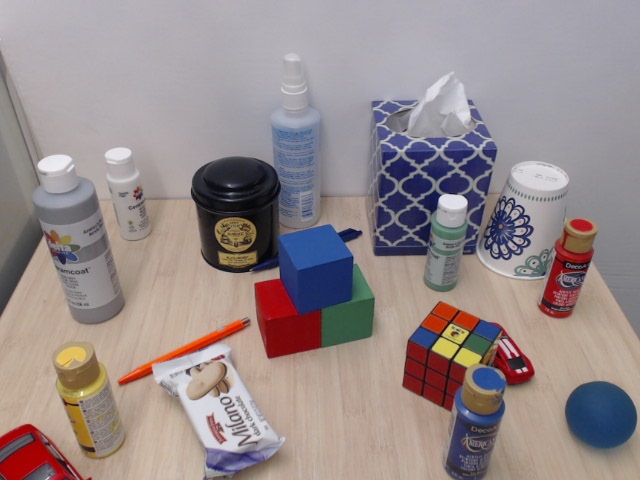} 
		}
		{
			\includegraphics[width=0.29\columnwidth]{fig/cubetest_00523.png} & 
			\includegraphics[width=0.29\columnwidth]{fig/cubetest_00290.png} & 
			\includegraphics[width=0.29\columnwidth]{fig/cubetest_01108.png} \\
			\includegraphics[width=0.29\columnwidth]{fig/cubetest_03637.png} &
			\includegraphics[width=0.29\columnwidth]{fig/cubetest_06148.png} &
			\includegraphics[width=0.29\columnwidth]{fig/cubetest_00415.png} 
		}		
	\end{tabular}
	\caption{One image from each of the six subsets of the data used in testing cube detection.  In lexicographic order:  isolated, cluttered, above table, occluded, stacked, and pyramid.}
	\label{fig:cubestackblueocc2}
\end{figure}

The dataset of $192$ images was split into six subsets, as illustrated in Fig.~\ref{fig:cubestackblueocc2}.  Images containing an isolated cube on a table, clutter on the table near the cube (including cubes of other colors), a cube elevated above the table by resting on another object, partial occlusion of a cube, inclusion in a stack of cubes, and inclusion in a pyramid of cubes.  The subsets consist of images with an equal number of annotations of red, green, blue, and yellow cubes, as well as a proportional number of stack sizes.

The results are shown in Table~\ref{tab:poseest_withrgby} where the false negative rate (FNR) is the percentage of cubes not detected.  Over all six scenarios, the MAE was $5.8$ pixels, and PCKh@$0.5$ was $90.0\%$.  In other words, $90.0\%$ of the vertices were detected within half the width of a square containing the same number of pixels as the convex hull of the ground truth cube.  Ignoring the $10\%$ outliers, the MAE was consistently $3.8$ pixels (that is, MAEc$=3.8$).  This result is remarkable, because all errors were measured with respect to the $400 \times 400$ image input to the network, whereas the final stages of the network output belief maps of size $50 \times 50$.  As a result, this means that the MAE of the inliers was actually less than $0.5$ pixels with respect to the belief maps.  Of the $219$ cubes in the dataset, $3$ were not detected at all by the network.  The rest of the numbers are with respect to the $216$ detected cubes.  Results are also fairly consistent across various bright colors, as shown in Table~\ref{tab:poseest_bycubecolor}, although we experienced more difficulty with black or white cubes.

\begin{table}%
\caption{Results of the perception network on the six scenarios.}
\begin{center}
\begin{tabular}{rcccc}
\toprule
 & MAE & PCKh@$0.5$ & MAEc@$0.5$ & FNR \\
 & (pixels) & (\%) & (pixels) & (\#/cubes) \\
\midrule
isolated    & $4.5$ & $90.6\%$ & $3.9$ & $0/20$ \\
cluttered   & $4.6$ & $90.9\%$ & $4.0$ & $2/24$ \\
above table & $4.0$ & $97.7\%$ & $3.7$ & $1/20$ \\
occluded    & $9.5$ & $83.6\%$ & $3.6$ & $0/24$ \\
stacked     & $6.4$ & $91.3\%$ & $3.7$ & $0/96$ \\
pyramid     & $4.9$ & $83.7\%$ & $3.8$ & $0/35$ \\
\textbf{all} & $\mathbf{5.8}$ & $\mathbf{90.0\%}$ & $\mathbf{3.8}$ & $\mathbf{3/219}$ \\
\bottomrule  
\end{tabular}
\end{center}
\label{tab:poseest_withrgby}
\end{table}

\begin{table}%
\caption{Results of the perception network by cube color.}
\begin{center}
\begin{tabular}{rcccc}
\toprule
 & MAE & PCKh@$0.5$ & MAEc@$0.5$ & FNR \\
 & (pixels) & (\%) & (pixels) & \#/cubes \\
\midrule
red    & $4.1$ & $92.1\%$ & $3.6$ & $2/55$ \\
green  & $4.8$ & $86.7\%$ & $4.0$ & $1/55$ \\
blue   & $5.8$ & $93.2\%$ & $3.6$ & $0/54$ \\
yellow & $8.4$ & $87.9\%$ & $3.8$ & $0/55$ \\
\bottomrule
\end{tabular}
\end{center}
\label{tab:poseest_bycubecolor}
\end{table}

\begin{figure}
	\centering
	\begin{tabular}{cc}
			\includegraphics[width=0.35\columnwidth,trim={0mm 0mm 0mm 3px},clip]{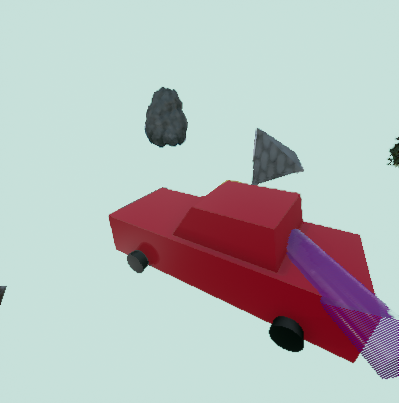} & 
	\ifthenelse{\boolean{useJPEG}}{
			\includegraphics[width=0.35\columnwidth,trim={1px 0mm 0mm 0mm},clip]{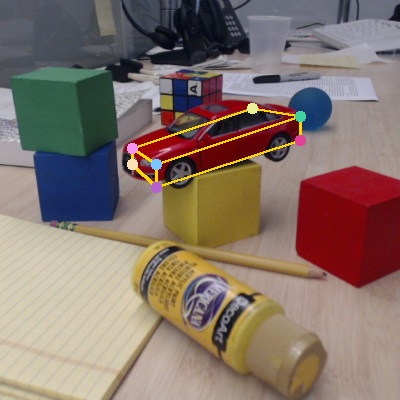} 
		}
		{
			\includegraphics[width=0.35\columnwidth,trim={1px 0mm 0mm 0mm},clip]{fig/00617.png} 
		}			
	\end{tabular}
	\caption{Example of detecting the bounding cuboid of an object that does not have a simple geometric shape.  Shown are a synthetic training image of a car (left), and a toy car detected in a real image (right) using the same method described for colored cubes.}
	\label{fig:cardetection}
\end{figure}

Although for practical reasons this work focuses on perceiving and manipulating solid cubes, the perception network as described is applicable to any rigid real-world object that can be reasonably approximated by its 3D bounding cuboid.  Shown in Fig.~\ref{fig:cardetection} is the output of the same network applied to a toy car, after training on simplistically simulated car images in the manner described above.  Unlike colored cubes, for which each face has identical appearance, objects such as cars have distinctive sides (front / back / left / right / top / bottom).  Although not obvious in the figure, the network is able to consistently maintain the identity of the vertices regardless of the relative roll, pitch, and yaw of the object with respect to the camera.

\subsection{Relationship inference}

\begin{table}
\caption{Relationship inference performance when trained with different augmentation methods.}
\label{tab:rel}
\begin{center}
\begin{tabular}{rcc}
\toprule
Augmentation                           & FPR              & FNR               \\
\midrule
None                                   & $1.4\%$          & $22.2\%$          \\
independent Gaussian ($\sigma\!=\!10^{-3}$) & $1.4\%$          & $22.2\%$          \\
structured Gaussian ($\sigma\!=\!5^{-4}$) & $1.6\%$          & $20.8\%$          \\
vertex confusion ($1\%$)               & $1.7\%$          & $19.4\%$          \\
vertex occlusion                       & $1.3\%$          & $13.9\%$          \\
all                                    & $1.9\%$          & $13.9\%$          \\
\bottomrule
\end{tabular}
\end{center}
\end{table}

We evaluated the performance of the relationship inference network using a hand-labeled dataset of scenes containing four cubes arranged using combinations of relationship types (\{{\sc Above}, {\sc Left}, {\sc Above-Pyramid}, {\sc None}\}), with some images containing multiple relationships and others none at all.
Across the $35$ images in the dataset, there were $72$ valid pairwise relationships among a total of $420$ cube pairs under evaluation by the network.
Cube vertices were detected using the cube detection networks discussed previously.

Results are shown in Table~\ref{tab:rel}. The false positive rate (FPR) indicates the rate at which non-adjacent cubes are predicted to be adjacent by the relationship network, while the false negative rate (FNR) indicates the rate at which valid relationships are missed.

The table evaluates four types of data augmentation, applied to the ground truth cube vertices in simulated training data: (1) independent Gaussian perturbation applied per vertex; (2) structured Gaussian perturbation applied per cube; (3) vertex confusion, whereby a vertex is randomly assigned to the location of another on the same cube; and (4) vertex occlusion, whereby vertices that overlap with a different cube are assigned a new random location with high variance.

Despite training only on simulated data, the relationship network transfers readily to the real world data. The best network achieves $\mbox{FPR}=1.3\%$ with $\mbox{FNR}=13.9\%$ when evaluated with a $0.5$ confidence threshold. The most effective augmentation (in fact, the only one that makes improvements to FNR and FPR) is simulated vertex occlusion. The other augmentations may be less necessary given the variation in viewing angles, distances, and random cube rotations and alignments already present in the simulated data, combined with the reliability of the detection network.

\subsection{Program generation \& execution}
\label{sec:experiment_program}

\begin{figure}
	\centering
		\includegraphics[width=0.8\columnwidth]{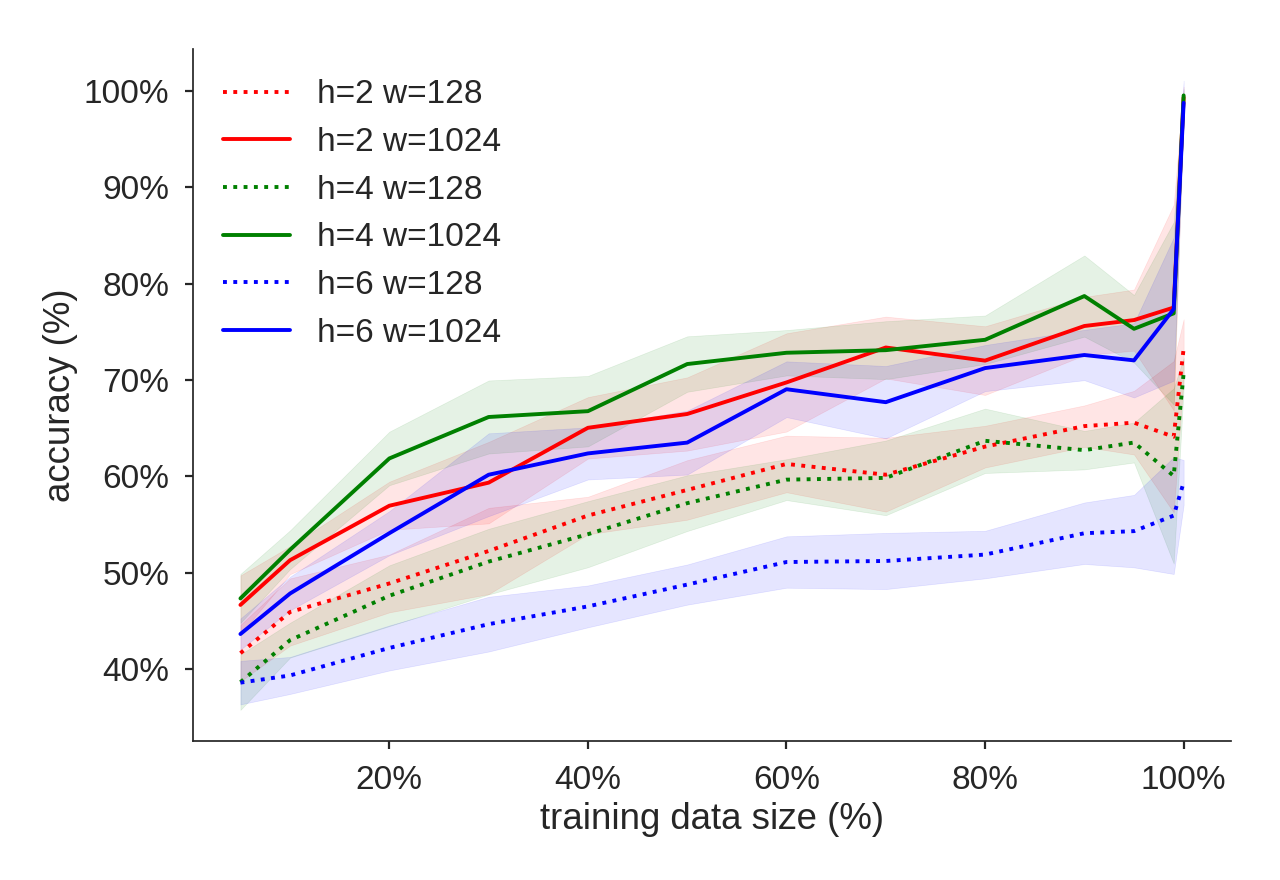}
	\caption{Comparison of various neural network architecture sizes for 
	predicting valid programs ($h$: number of hidden layers, $w$: number of units per hidden layer).   The $x$-axis represents the percentage of data 
	used during training (ranging from $1\%$ to $95\%$), whereas the $y$-axis is the accuracy on the entire dataset.
	The best results are achieved with $h=4$, $w=1024$ (solid green line).}
	\label{fig:perf_plan}
\end{figure}

The generalization performance of the program generation network was evaluated by testing its accuracy with respect to the percentage of all possible stacks used for training data.\footnote{Since the percepts are discrete, all possible stacks can be enumerated.}
Fig.~\ref{fig:perf_plan} shows the accuracy (number of times the network 
predicted the correct pick or place) versus training data size %
for several different networks.  
The smallest network (red dotted line), with $2$ hidden layers and $128$ units per hidden layer, is not even able to memorize the dataset (since it achieves just $83\%$ accuracy even with $95\%$ training data).  The largest network (blue solid line), with $6$ hidden layers and $1024$ units per hidden layer, is not only able to memorize the dataset but also achieves reasonable generalization performance of $71\%$ when trained on $80\%$ of the dataset.  Since $w=512$ (not shown to avoid clutter) achieves similar results to $w=128$, we conclude that 1024 units are necessary.  However, although the number of hidden units is crucial to good performance, the number of hidden layers is less so; in fact, 4 layers outperforms 6.   

\begin{figure}
	\centering
	\ifthenelse{\boolean{useJPEG}}{
			\includegraphics[width=0.4\columnwidth]{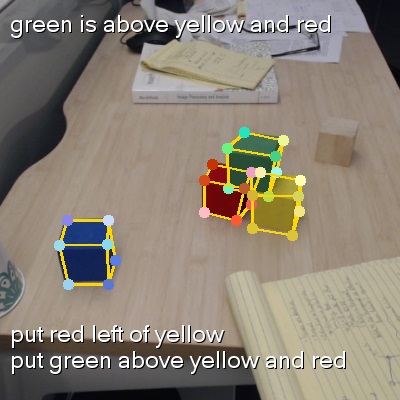}
		}
		{
			\includegraphics[width=0.4\columnwidth]{fig/plan_example.png}
		}			
	\caption{The correct program (text at the bottom of the image) is learned in this case, even though the state is ambiguous (text at the top of the image) because it is missing the {\sc left} 
 relationship.}
	\label{fig:plan_pyramid}
\end{figure}

Fig.~\ref{fig:plan_pyramid} shows the program learned by the network to reproduce a three-cube pyramid.  What is interesting about this case is that the state is ambiguous since the relationship network failed to predict that the red cube is left of the yellow cube, perhaps due to the large gap between the two cubes.
Nevertheless, the network has learned to infer that there should be a third relationship in the pyramid and returns a program that first creates a left relationship, before indicating how the green cube should be stacked.
Of course, in this case the network does not have enough information to decide which of the two symmetric left relationships (``red left of yellow'' or ``yellow left of red'') is correct, since the coordinates of the cubes are not available. But the network's guess indicates behavior that is more than mere memorization of the training data.

The execution network was tested in a similar manner by measuring the influence of withholding training data. 
Table~\ref{tab:exec} presents the accuracy versus the amount of data withheld. 
The data consisted of approximately $36$k data points from all possible arrangements of 
$2$- to $6$-cube stacks.  Each result is the average over the five last epochs from five independently trained networks. When the network is trained on the whole collection, it achieves perfect accuracy, as expected, but it also works surprisingly well when only half the data is present for training, thus demonstrating generalization.  

\begin{table}
\caption{Execution network performance (on entire dataset) as a function of the amount of data used for training.}
\label{tab:exec}
\begin{center}
\begin{tabular}{rcccc}
\toprule
training data & $5\%$  & $10\%$ & $50\%$ & $95\%$ \\
\midrule
accuracy      & $56\%$ & $85\%$ & $97\%$ & $98\%$ \\
\bottomrule
\end{tabular}
\end{center}
\end{table}

\subsection{Entire system}

The networks were connected and tested in our lab with a Baxter robot.  For perception, we used an Intel RealSense SR300 camera, whose exterior orientation with respect to the robot was determined using a standard calibration approach.  RGB images were used for perception, while depth images were used to determine the 3D position of the cubes with respect to the robot.

\begin{figure*}
	\centering
	\begin{tabular}{cccc}
	\ifthenelse{\boolean{useJPEG}}{
			\includegraphics[width=0.45\columnwidth]{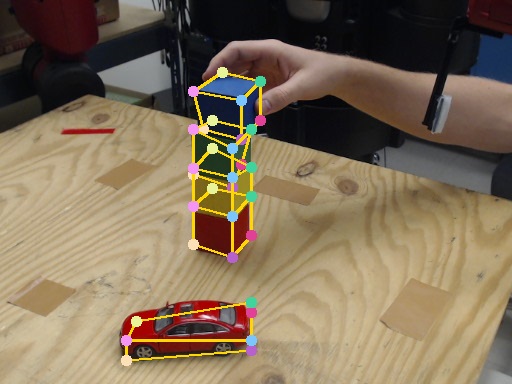} &
			\includegraphics[width=0.45\columnwidth]{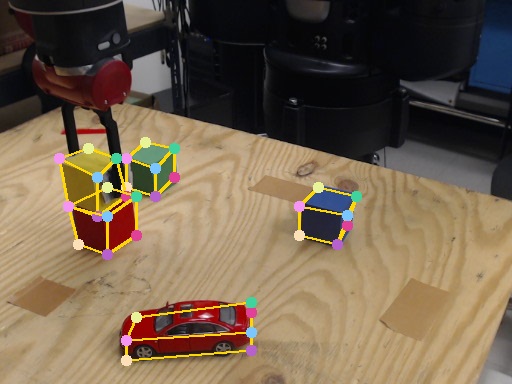} &
			\includegraphics[width=0.45\columnwidth]{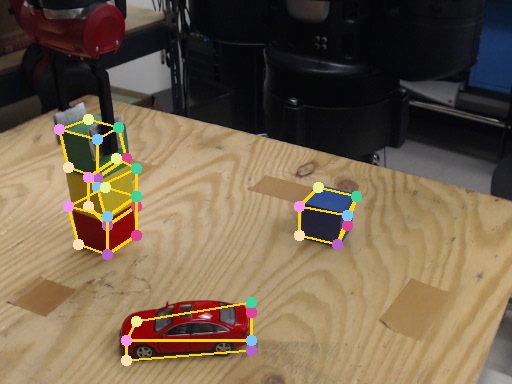} &
			\includegraphics[width=0.45\columnwidth]{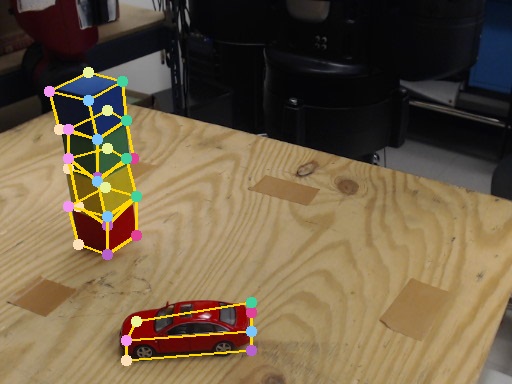} \\
			\includegraphics[width=0.45\columnwidth]{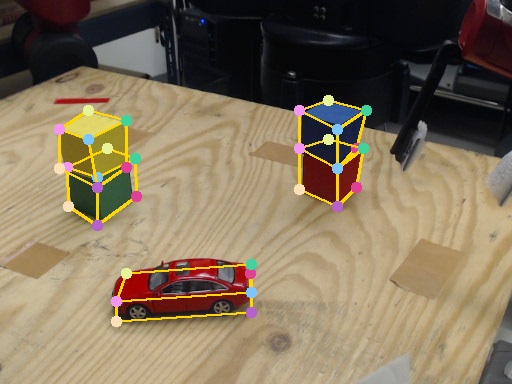} &
			\includegraphics[width=0.45\columnwidth]{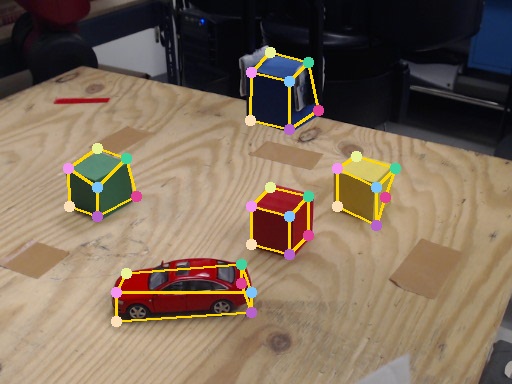} &
			\includegraphics[width=0.45\columnwidth]{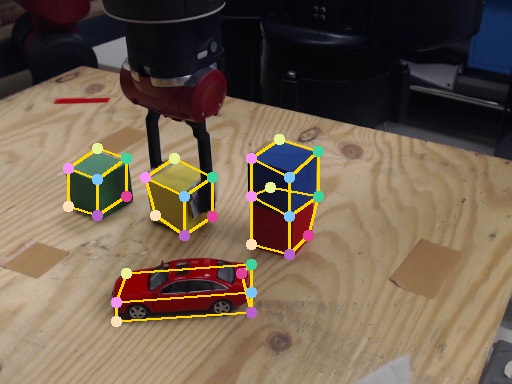} &
			\includegraphics[width=0.45\columnwidth]{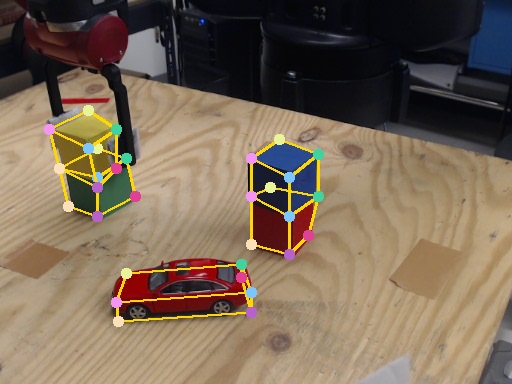} \\
			\includegraphics[width=0.45\columnwidth]{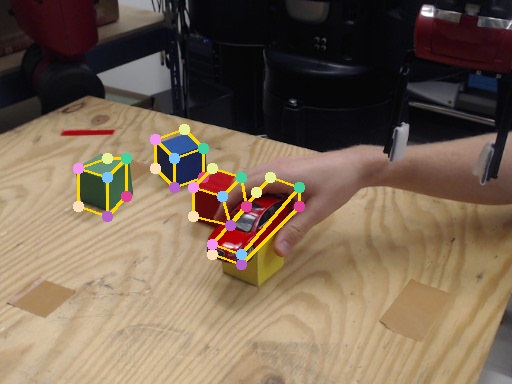} &
			\includegraphics[width=0.45\columnwidth]{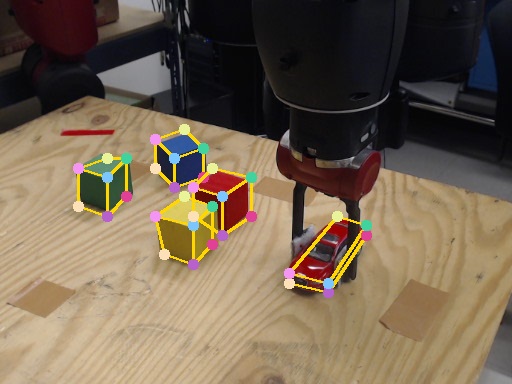} &
			\includegraphics[width=0.45\columnwidth]{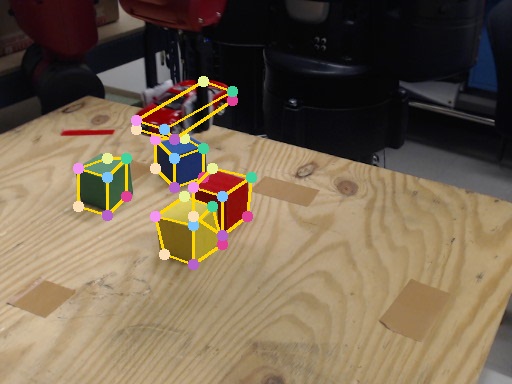} &
			\includegraphics[width=0.45\columnwidth]{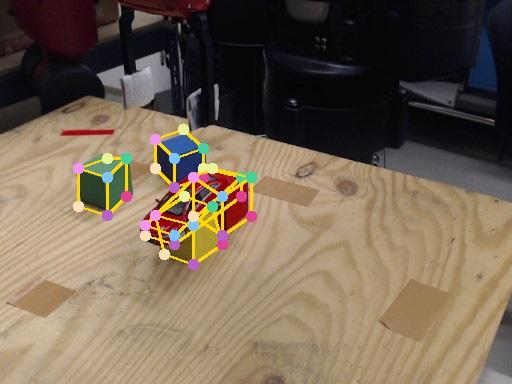} \\
		}
		{
			\includegraphics[width=0.45\columnwidth]{fig/demo_4cube/00000020_crop.png} &
			\includegraphics[width=0.45\columnwidth]{fig/demo_4cube/00000079_crop.png} &
			\includegraphics[width=0.45\columnwidth]{fig/demo_4cube/00000119_crop.png} &
			\includegraphics[width=0.45\columnwidth]{fig/demo_4cube/00000165_crop.png} \\
			\includegraphics[width=0.45\columnwidth]{fig/demo_2stacks/00000015_crop.png} &
			\includegraphics[width=0.45\columnwidth]{fig/demo_2stacks/00000088_crop.png} &
			\includegraphics[width=0.45\columnwidth]{fig/demo_2stacks/00000126_crop.png} &
			\includegraphics[width=0.45\columnwidth]{fig/demo_2stacks/00000174_crop.png} \\
			\includegraphics[width=0.45\columnwidth]{fig/car_stacking/00000017_crop.png} &
			\includegraphics[width=0.45\columnwidth]{fig/car_stacking/00000058_crop.png} &
			\includegraphics[width=0.45\columnwidth]{fig/car_stacking/00000062_crop.png} &
			\includegraphics[width=0.45\columnwidth]{fig/car_stacking/00000071_crop.png} \\
		}			
		demonstration &  \multicolumn{2}{c}{execution} & final
		\end{tabular}
	  \caption{Three demonstrations (left column), and snapshots of the robot executing the automatically generated program (remaining columns).  From top to bottom, the programs are as follows:  ``Place yellow on red, green on yellow, and blue on green'' (top row), ``Place yellow on green, and blue on red'' (middle row), and ``Place car on yellow'' (bottom row).  In the second row, notice that the robot recovered from an error in initially misplacing the yellow cube.}
	\label{fig:resultsRobot}
\end{figure*}

To test the system, a human stacked the cubes together, then pressed a button to indicate that the demonstration was complete.  The system automatically inferred the positions of the cubes, their relationships, and the human-readable program.  After the human rearranged the cubes and pressed another button, the robot then performed the pick-and-place steps output by the execution network.  In addition to cubes, we also performed experiments involving a toy car (mentioned earlier) placed on a cube.  Some examples are shown in Fig.~\ref{fig:resultsRobot}.

Qualitatively, the system worked fairly reliably, but a few limitations are worth mentioning.  First, in our setup the intersection of the robot arm's working area and the camera's field of view yielded an effective working area approximately $20$~cm by $30$~cm, which was rather small.  Another challenge involved detecting the severely occluded cubes at the bottom of the stack; coupled with the fact that the execution network does not retain any memory of its previous actions, any such missed detections led to incorrect behavior in the execution.  Obviously, open loop execution would solve this problem but at the expense of limiting generalizability of the system.  Moreover, the open gripper was only 9~cm wide, leading to just a 2~cm tolerance for grasping the 5~cm cubes, which was close to the accuracy of our detection system; as a result, the gripper would occasionally fail to envelope the intended cube.  Finally, even though all cubes were simply painted wooden blocks, the depth sensor of the RealSense camera in our lab struggled inexplicably to obtain meaningful measurements of the green cube at certain angles, thus further challenging the system.

\section{CONCLUSION}
\label{sec:conclusion}

We have presented a system to generate human-readable programs from a real-world demonstration.  The system consists of a sequence of neural networks to perform tasks associated with perception, program generation, and program execution.  For perception, we introduce image-centric domain randomization leveraging convolutional pose machines, which results in a vision-based network that can be applied to any camera, without assumptions about the camera pose or the presence of specific background features in the scene.  For program generation and execution, we show that fully connected networks, despite their simplicity, generalize surprisingly well when considering relationships, states, and programs not encountered during training.  Although training individual networks separately may be suboptimal compared with end-to-end training, it facilitates component-wise testing, interpretability, and modularity.  There remain many issues to explore, such as increasing robustness of domain randomization to variations in lighting and color, incorporating context to better handle occlusion, leveraging past execution information to overcome limitations in sensing, and expanding the vocabulary of the human-readable programs.

\section*{ACKNOWLEDGMENTS}

The authors would like to thank the reviewers for their helpful comments.  Appreciation also goes to
Omer Shapira, Mark Brophy, Hai Loc Lu, 
and Bryan Dudesh for sharing their expertise and insights regarding the design 
and implementation of the simulator used for domain randomization, as well as for many helpful discussions.

\vfill


\begin{thebibliography}{99}
\bibitem{abbeel04}
  P.~Abbeel and A.~Y. Ng.
  Apprenticeship learning via inverse reinforcement learning, in \emph{ICML}, July 2004.
\bibitem{andriluka2014cvpr} 
  %
  M. Andriluka et al.
  2D human pose estimation: New benchmark and state of the art analysis, in \emph{CVPR}, June 2014.
\bibitem{cao2016arx} 
  Z. Cao, T. Simon, S.-E. Wei, Y. Sheikh.
  Realtime multi-person 2D pose estimation using part affinity fields, in \emph{CVPR}, June 2017.
  %
%
%
%
%
	%
\bibitem{deng2009in}  
  J. Deng, W. Dong, R. Socher, L.-J. Li, K. Li, L. Fei-Fei.
  ImageNet: A large-scale hierarchical image database, in \emph{CVPR}, June 2009.
\bibitem{denil17a}
M.~Denil, S.~G. Colmenarejo, S.~Cabi, D.~Saxton, and N.~de~Freitas.
  Programmable agents, arXiv preprint, arXiv:1706.06383, 2017.
	%
\bibitem{devin2017icra}
  %
  C. Devin et al.
  Learning modular neural network policies for multi-task and multi-robot transfer, in \emph{ICRA}, 2017.
\bibitem{duan2017arx}  
  %
  Y. Duan et al.  
	One-shot imitation learning, in \emph{NIPS}, Dec.~2017.
	%
\bibitem{feniello2014iros} 
  A. Feniello, H. Dang, and S. Birchfield.  
	Program synthesis by examples for object repositioning tasks, in \emph{IROS}, Sept.~2014.
\bibitem{ferrari2008cvpr}  
  V. Ferrari, M. Mar{\'i}n-Jim{\'e}nez, and Andrew Zisserman.
	Progressive search space reduction for human pose estimation, in \emph{CVPR}, 2008.
\bibitem{finn16}
  C.~Finn, S.~Levine, and P.~Abbeel.
	Guided cost learning: Deep inverse optimal control via policy optimization, in \emph{ICML}, June 2016. 
	%
\bibitem{garnello16a}
  M.~Garnelo, K.~Arulkumaran, and M.~Shanahan.
	Towards deep symbolic reinforcement learning, in \emph{NIPS Deep RL Workshop}, 2016.
	%
	%
\bibitem{gupta2017arx} 
  %
  T. Gupta et al.
  Aligned image-word representations improve inductive transfer across vision-language tasks, in \emph{ICCV} 2017.
	%
\bibitem{hausmann17}
  %
  K.~Hausman et al.
  Multi-modal imitation learning from unstructured demonstrations using generative adversarial nets, in \emph{NIPS}, 2017.
	%
	%
\bibitem{ho16}
  J.~Ho and S.~Ermon.
  Generative adversarial imitation learning, in \emph{NIPS}, 2016.
	%
  %
\bibitem{Hu00}
  %
  H.~H. Hu et al. 
	Visual cues for imminent object contact in realistic virtual environment, in \emph{Proc. of the Conf. on Visualization}, 2000.
\bibitem{james2017corl}
 %
  S. James et al.
  Transferring end-to-end visiomotor control from simulation to real world for a multi-stage task, in \emph{CoRL}, 2017.
\bibitem{johnson2017cvpr}
%
  J. Johnson et al.
CLEVR: A diagnostic dataset for compositional language and elementary visual reasoning, in \emph{CVPR}, June 2017.
\bibitem{karpathy2015cvpr} A. Karpathy, L. Fei-Fei.
  Deep visual-semantic alignments for generating image descriptions, in \emph{CVPR}, June 2015.
\bibitem{ng00irl}
  A.~Y. Ng and S.~J. Russell.
	Algorithms for inverse reinforcement learning, in \emph{ICML}, June 2000.
\bibitem{paul2017ijcai} 
  %
  R. Paul et al.
  Temporal grounding graphs for language understanding with accrued visual-linguistic context, in \emph{IJCAI}, 2017.
\bibitem{peyre2017arx} 
  J. Peyre, I. Laptev, C. Schmid, J. Sivic. 
  Weakly-supervised learning of visual relations, in \emph{ICCV}, 2017.
	%
\bibitem{pomerleau91}
  D.~Pomerleau. 
	Efficient training of artificial neural networks for autonomous navigation, \emph{Neural Computation}, 3(1):88--97, 1991.
  %
\bibitem{ross10a}
  S.~Ross and D.~Bagnell.
  Efficient reductions for imitation learning, in \emph{AISTATS}, pp. 661--668, 2010. 
	%
	%
\bibitem{ross11a}
  %
  S.~Ross et al.
	A reduction of imitation learning and structured prediction to no-regret online learning, in \emph{AISTATS}, April 2011.
  %
\bibitem{shen2017arx} 
  Z. Shen, J. Li, Z. Su, M. Li, Y. Chen, Y.-G. Jiang, X. Xue. 
  Weakly supervised dense video captioning, in \emph{CVPR}, June 2017.
	%
\bibitem{simonyan2015iclr} 	
  K. Simonyan and A. Zisserman.  
	Very deep convolutional networks for large-scale image recognition, in \emph{ICLR}, 2015.
\bibitem{tobin2017iros}  
  J. Tobin, R. Fong, A. Ray, J. Schneider, W. Zaremba, P. Abbeel.  
	Domain randomization for transferring deep neural networks from simulation to the real world, in \emph{IROS}, 2017.
%
\bibitem{vinyals2015cvpr} O. Vinyals, A. Toshev, S. Bengio, D. Erhan.
Show and tell: A neural image caption generator, in \emph{CVPR}, June 2015.
\bibitem{wei2016arx} S.-E. Wei, V. Ramakrishna, T. Kanade, Y. Sheikh.
Convolutional pose machines, in \emph{CVPR}, 2016.
%
\bibitem{winograd1971thesis} T. Winograd. 
  Procedures as a representation for data in a computer program for understanding natural language, MIT MAC-TR-84, 1971.
\bibitem{xu2017:ntp}
  D. Xu, S. Nair, Y. Zhu, J. Gao, A. Garg, L. Fei-Fei, S. Savarese.
	Neural task programming: Learning to generalize across hierarchical tasks, in arXiv 1710.01813, 2017.
\bibitem{yang2013pami} Y. Yang and D. Ramanan. Articulated human detection with flexible mixtures-of-parts. \emph{PAMI}, 35(12):2878--2890, Dec. 2013.
%
%
%
  %
  %
	%
\end{thebibliography}
\end{document}